\documentclass{article}

\PassOptionsToPackage{numbers, compress}{natbib}

\usepackage[final]{neurips_2022}

\usepackage[utf8]{inputenc} 
\usepackage[T1]{fontenc}    
\usepackage{hyperref}       
\usepackage{url}            
\usepackage{booktabs}       
\usepackage{amsfonts}       
\usepackage{nicefrac}       
\usepackage{microtype}      
\usepackage{xcolor}         
\usepackage{graphicx}
\usepackage{caption}
\usepackage{subcaption}

\title{Exploring the Performance of Pruning Methods in Neural Networks: An Empirical Study of the Lottery Ticket Hypothesis}

\author{%
   Eirik Fladmark \\
   University of Cambridge \\
   \texttt{ef454@cam.ac.uk} \\
   \And
   Laura B. Justesen \\
   University of Cambridge \\
   \texttt{lbj25@cam.ac.uk} \\
   \AND
   Muhammad H. Sajjad \\
   University of Cambridge \\
   \texttt{mhs57@cam.ac.uk} \\
}

\begin{document}

\maketitle

\begin{abstract}
In this paper, we explore the performance of different pruning methods in the context of the lottery ticket hypothesis. We compare the performance of L1 unstructured pruning, Fisher pruning, and random pruning on different network architectures and pruning scenarios. The experiments include an evaluation of one-shot and iterative pruning, an examination of weight movement in the network during pruning, a comparison of the pruning methods on networks of varying widths, and an analysis of the performance of the methods when the network becomes very sparse. Additionally, we propose and evaluate a new method for efficient computation of Fisher pruning, known as batched Fisher pruning. 
\end{abstract}

\section{Introduction}

\subsection{Pruning}

For more than 30 years, pruning has been used to reduce the size of neural networks. For example, Optimal Brain Damage (OBD) is a pruning technique used by LeCun et al. \cite{lecun:brain-damage} to prune unimportant weights from a neural network, which led to a significant improvement in speed and a small increase in accuracy. Furthermore, pruning can improve generalisation since smaller networks are less prone to over-fitting \cite{Hagiwara1994}.

\subsection{Pruning methods}

There exist many different pruning algorithms that can be applied to a neural network, such as penalty-based pruning \cite{Nielsen2008}, mutual information pruning \cite{Xing2009}, evolutionary pruning \cite{whitley1990}, sensitivity-based pruning \cite{lecun:brain-damage}, and magnitude-based pruning \cite{Augasta2013}.

\textbf{Random pruning:} Random pruning is probably the simplest and most naive way to prune a network \cite{Liu2022}. Here a certain percentage of the network weights are pruned at random. Thus, it does not attempt to determine whether some weights matter more than others. As a result, it usually leads to poor performance.

\textbf{Magnitude-based pruning:} A relatively simple type of pruning algorithm is magnitude-based pruning, which is based on the assumption that small weights are irrelevant \cite{Augasta2013}. It therefore prunes small weights from the network. These algorithms are usually easy to compute and thus quite time-efficient \cite{Augasta2013}. However, while Sietsma \& Dow \cite{Sietsma1988} found that the removal of redundant weights led to improved performance, further pruning resulted in degraded performance on noisy data. Thus, some level of redundancy might be beneficial.

\textbf{L1 pruning:} An example of magnitude-based pruning is L1 pruning where the weights with the smallest L1 norm are pruned (see Equation \ref{eq:l1-norm} taken from \cite{wiki-norm}).

\begin{equation} \label{eq:l1-norm}
    ||\textbf{x}||_1 = \sum^{n}_{i=1} |x_i|
\end{equation}

\textbf{Other categorisations:} Furthermore, pruning algorithms can be classified according to two additional dimensions: structured--unstructured and global--local \cite{gavrikov2022}. In structured pruning, individual parameters are pruned without considering their structure, whilst structured pruning involves the removal of whole structures of parameters (e.g., convolutional filters). In local pruning, pruning is performed per layer in the network, whilst global pruning is applied over all layers. In this paper, we only worked with global pruning methods.

\textbf{Fisher pruning:} Theis et al. \cite{fisher-pruning} introduced Fisher pruning and showed that it can be used to significantly reduce the computational complexity of state-of-the-art gaze models while maintaining similar performance. Fisher pruning is based on the concept of Fisher information, which is used to estimate the increase in loss when removing a specific weight from the network. This approximation is used to select weights for pruning where Fisher pruning removes the weights with the smallest expected increase in loss. Theis et al. \cite{fisher-pruning} implemented pruning of individual weights as well as pruning of entire feature maps (especially suited for convolutional networks). In this study, we focus on the pruning of individual weights. In this case, the expected increase in loss when pruning the \textit{k}th parameter can be calculated according to Equation \ref{eq:fisher-pruning} (equation taken from \cite{fisher-pruning}):

\begin{equation} \label{eq:fisher-pruning}
    \Delta_k = \frac{1}{2N} \theta^2_k \sum_{n=1}^{N}g^2_{nk}
\end{equation}

where \textit{N} is the number of data points used to estimate the Fisher information, \(\theta_k\) is the \textit{k}th parameter, and \(g_{nk}\) is the gradient of the \textit{k}th parameter with respect to the \textit{n}th data point. 

\subsection{Lottery Ticket Hypothesis}

\textbf{Definition:} The lottery ticket hypothesis was introduced by Frankle \& Carbin \cite{Frankle} and states the following: \textit{Given a dense neural network that has been trained from randomly initialised weights, there exists a sub-network, which (when trained in isolation) will achieve at least the same test accuracy as the original network when trained for at most the same number of iterations}. Such a trainable sub-network is referred to as a winning ticket.

\textbf{Finding winning tickets:} Winning tickets can be found using standard pruning methods. Since pruning aims to remove unimportant weights, it is possible to apply pruning to the trained network to find suitable sub-networks. In this study, we consider two approaches to using pruning to find winning tickets: one-shot and iterative \cite{Frankle}. One-shot pruning is an approach where the network is trained before ${p\%}$ of the weights in the network are pruned away. The weights are then reset to their initial state, and the network is trained until convergence again. However, iterative pruning trains a network, prunes $p^{\frac{1}{n}}\%$ of the current weights, and resets the weights to their initial state. It then repeats this process until a total of $p\%$ of the weights are pruned, at which point it trains the network till convergence one last time. It has been found that iterative pruning is better at finding winning tickets when networks become increasingly sparse \cite{Frankle}. However, this comes at a cost, as it is more computationally expensive than one-shot pruning.

\textbf{Recent research:} Since the lottery ticket hypothesis was first introduced \cite{Frankle}, a lot of research has been conducted in this field. Here we will summarise some relevant studies to show the range of research directions pursued.

One area of interest is the timing of the pruning operation. Initially, Frankle et al. \cite{Frankle} focused on pruning after the network had been trained. In a later study \cite{Frankle-6}, they compared the performance of winning tickets based on whether the pruning occurred at initialisation or early in training (between 0.1\% and 7\% through training). They noted that pruning at initialisation generally fails for deeper networks. However, when pruning is moved from initialisation to early in training, they were able to find small sub-networks that, after completing the rest of the training, matched the performance of the original network.

Frankle et al. \cite{Frankle-3} further extended these findings by applying instability analysis to the lottery ticket hypothesis. They found that sub-networks identified through iterative magnitude pruning are only successful if they are stable to the noise in stochastic gradient descent.

Another area of research is the resetting scheme used after pruning. In the original paper on the lottery ticket hypothesis \cite{Frankle}, the weights of the winning ticket were reset to the original initialisations. However, Frankle et al. \cite{Frankle-2} subsequently introduced \textit{late resetting} where the weights are reset to the weights obtained after training for a few iterations. This allowed them to find successful winning tickets for a large state-of-the-art network (ResNet-50).

Sub-networks identified by the lottery ticket hypothesis have also been found to facilitate transfer learning. Chen et al. \cite{Chen2021} investigated the application of the lottery ticket hypothesis to supervised and self-supervised pre-trained models for computer vision tasks. Interestingly, it was found that the winning tickets identified for pre-trained networks also applied to downstream tasks. In fact, with 59.04\% to 96.48\% pruning, the sub-networks continued to perform well without any degradation on several downstream tasks (classification, detection, and segmentation).

Finally, different approaches to retraining the pruned network have been explored. One such retraining technique is \textit{fine-tuning} where a small fixed learning rate is used used to train the unpruned weights from their final trained values \cite{Renda}. Another retraining technique, proposed by Frankle et al. \cite{Frankle-3} is \textit{weight rewinding} where the unpruned weights are rewinded to an earlier point in training and then retrained from there with the original training schedule. Renda et al. \cite{Renda} proposed a third retraining technique, \textit{learning rate rewinding}, where the unpruned weights are trained from their final values with the learning rate schedule from weight rewinding. Renda et al. \cite{Renda} found that weight rewinding and learning rate rewinding outperformed fine-tuning. Based on these results, they argued that the rewinding techniques could be used in a network-agnostic pruning algorithm that would be able to match more network-specific techniques.

\subsection{Project introduction}

This study is influenced by the work of Frankle \& Carbin \cite{Frankle} and Theis et al. \cite{fisher-pruning}.

\textbf{Motivation:} There are several reasons for considering the lottery ticket hypothesis as an interesting field of research. It suggests that small neural networks (such as the winning tickets) are capable of learning as effectively as larger networks (i.e., obtain similar performance when trained for the same number of iterations) \cite{Frankle-2}. Given their small size, such networks can be trained much faster than their larger counterparts. Thus, insights from research on the lottery ticket hypothesis could potentially lead to more computationally efficient training procedures (e.g., by identifying suitable sub-networks early in training). Furthermore, research on the lottery ticket hypothesis can inspire the development of new and better initialisation schemes and architectures \cite{Frankle}. Lastly, smaller network sizes can improve the interpretability of machine learning models. Since the performance of the winning ticket is affected by the pruning method used, it is worthwhile to study the performance of different pruning methods.

\textbf{Contributions:}

\begin{itemize}
    \item We reproduce selected experiments from \cite{Frankle}.
    \item We implement iterative Fisher pruning for the identification of winning tickets and compare it to other pruning methods.
    \item We perform further experimentation to investigate the performance of the different pruning methods under various conditions.
    \item We introduce batched Fisher pruning, which extends the original Fisher pruning algorithm to work on batches, and investigate the effect of different batch sizes.
\end{itemize}

\section{Implementation details}

For the experiments conducted in this report, we used the OpenLTH framework developed by Facebook Research \cite{open_lth}. This framework implements some of the latest research on the lottery ticket hypothesis (such as \cite{Frankle, Frankle-2, Frankle-3, Frankle-4, Frankle-5}) and was therefore an essential tool that allowed us to quickly start experimenting. Furthermore, it was simple to use, and it made it easy to implement new features as required for the different experiments. Lastly, the code and implementation details of our adaptation of Fisher pruning can be found here: \url{https://github.com/Fladmark/open_lth/blob/main/pruning/fisher_global.py}. 

\section{Experiments}

\subsection{One-shot pruning versus iterative pruning}

\begin{figure}[t]
\centering
\includegraphics[width=0.8\textwidth]{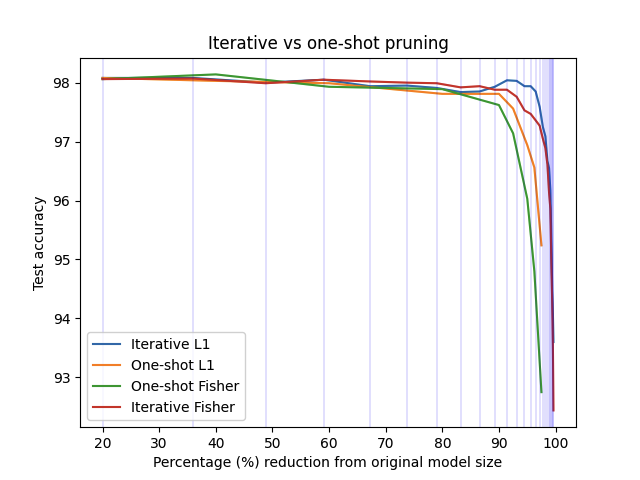}
\caption{Comparison of test accuracy for one-shot pruning and iterative pruning. The vertical lines mark the locations of the pruning iterations in iterative pruning.}
\label{fig:one-shot}
\end{figure}

Our base implementation of the lottery ticket hypothesis is based on the iterative pruning paradigm, where the network is trained and pruned through several iterations until the desired reduction in network size has been achieved. An alternative to this approach is one-shot pruning where the network is trained once after which the network size is reduced in a single pruning step \cite{Frankle}. Both approaches have advantages and disadvantages. We expect iterative pruning to yield winning tickets with higher accuracy than one-shot pruning. However, the repeated training iterations make iterative pruning relatively expensive. On the other hand, since one-shot pruning only trains (and prunes) the network once, it is much cheaper to use.

To test this, we compared iterative and one-shot pruning for L1 unstructured pruning and Fisher pruning. Figure \ref{fig:one-shot} shows how the test accuracy changes with the percentage of weights pruned. Surprisingly, we observe that one-shot pruning and iterative pruning have very similar performances up to around 50\% reduction. After this point, performance differences become noticeable with the iterative pruning methods resulting in slightly higher accuracies. The largest performance differences are seen when pruning exceeds 90\%. Here we observe that the one-shot pruning methods exhibit steep drops in accuracy before the iterative methods. In general, at high levels of pruning, iterative pruning outperforms one-shot pruning in terms of accuracy. Furthermore, within each pruning paradigm, L1 unstructured pruning results in higher accuracy than Fisher pruning. However, for smaller reductions in network size, one-shot pruning results in accuracy performance similar to iterative pruning. Thus, in cases where speed is essential and medium pruning is desired, one-shot pruning might be the best choice.

\subsection{Pruning and weight movement}

An interesting metric of comparison between different pruning methods is the movement in weight space. The goal of pruning is to remove "unnecessary" weights from the network and only retain the most relevant weights. During iterative pruning, weights are repeatedly trained and updated. They are therefore likely to move away from their initial value. In this experiment, we want to investigate whether the choice of pruning method affects the amount of movement in weight space. The weight movement is calculated according to Equations \ref{eq:weight-abs} and \ref{eq:weight-avg}.

\begin{equation} \label{eq:weight-abs}
    weight_{abs\_dif\_i} = sum(abs(W_{0\_m}-W_{i\_m}))
\end{equation}

\begin{equation} \label{eq:weight-avg}
    weight_{avg\_dif\_i} = \frac{weight_{abs\_dif\_i}}{N_i}
\end{equation}

where \(W_{0\_m}\) is the weight matrix of the initial unpruned network after training for \textit{m} epochs, \(W_{i\_m}\) is the weight matrix of the network after \textit{i} pruning iterations trained for \textit{m} epochs, and \(N_i\) is the number of unpruned weights after \textit{i} pruning iterations. \(abs()\) takes a matrix as input and returns a matrix where each element is the absolute value of the corresponding element in the input matrix. \(sum()\) takes a matrix as input and returns the sum of its elements. It should be noted that when we compute the subtraction in Equation \ref{eq:weight-abs}, we only use the weights in \(W_{0\_m}\) that have not been pruned after \textit{i} iterations.

\begin{figure}[h]
\centering
\includegraphics[width=0.8\textwidth]{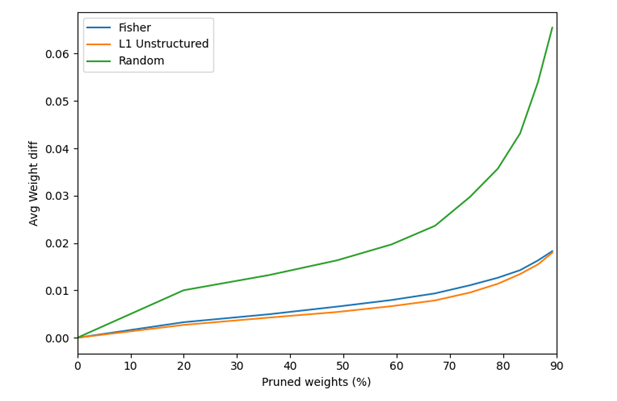}
\caption{Average weight movement for different pruning methods (\textit{m} = 10 epochs).}
\label{fig:weight-movement}
\end{figure}

Figure \ref{fig:weight-movement} shows the average weight movement for the network at different percentages of pruning. We observe that as the percentage of pruned weights increases, the average weight movement increases as well. This could be because individual weights in a highly pruned network have to move more (on average) to get into a suitable configuration. In a larger network, this collective movement is spread over more weights, which might lead to smaller average weight movement (per weight). Furthermore, we note that random pruning results in higher weight movement than the other pruning methods. Since random pruning removes weights without reference to any measure of relevance, it might end up pruning important weights. When important weights are removed, the remaining weights will have to adapt to compensate for their loss. This can potentially explain the significantly higher weight movement for random pruning.

In Figure \ref{fig:weight-movement}, Fisher pruning and L1 unstructured pruning have very similar average weight differences. However, Fisher pruning has a slightly higher weight difference throughout. Fisher pruning removes weights based on the Fisher information. This calculation (see Equation \ref{eq:fisher-pruning}) is based on gradients, which means that a parameter associated with smaller gradients has a higher chance of being pruned. We theorise that Fisher pruning is therefore likely to prune weights with smaller updates, which means that weights with (comparatively) larger updates are left unpruned. Hence, after pruning, the average weight difference would increase. Based on this, we expect Fisher pruning to have higher weight movement than L1 unstructured pruning, but less than random pruning. However, we did not expect Fisher pruning to be as close to L1 unstructured pruning as occurs in Figure \ref{fig:weight-movement}. We expected to see a bigger weight difference for Fisher pruning---instead, it remains quite close to L1 unstructured pruning. This might suggest that pruning weights with a gradient-based measure does not significantly affect the amount to which the weights move during pruning.

\subsection{Varying network width} \label{section:width}

\begin{figure}[h]
\centering
\includegraphics[width=0.8\textwidth]{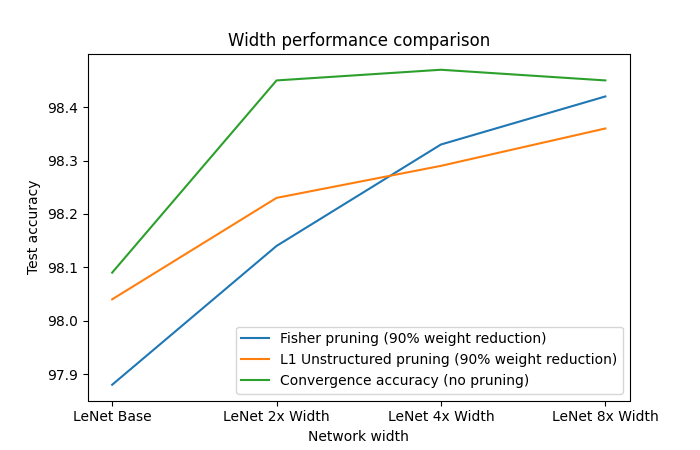}
\caption{Accuracy of different models (pruned and unpruned) for varying network widths.}
\label{fig:width-comparson}
\end{figure}

It is hard to conclude which pruning method is better than others, especially when only considering networks of a fixed size. A pruning method that works well for medium-sized networks might be outperformed by another pruning method when applied to larger networks. It is therefore desirable to compare performance for a range of different network sizes. In this experiment, we compare Fisher pruning and L1 unstructured pruning on networks of varying widths. We start with a LeNet with 2 hidden layers of size 300 and 100 respectively. Three scaled-up versions of the network are then tested: (600, 200), (1200, 400), and (2400, 800) neurons per hidden layer. For each network, we iteratively prune 20\% of the weights 10 times, which results in the network size being reduced by approximately 90\%. The results are shown in Figure \ref{fig:width-comparson}.

Figure \ref{fig:width-comparson} illustrates the difficulty of identifying an overall best pruning method since neither pruning method dominates the other. It seems that Fisher pruning performs better on wider networks, whilst L1 unstructured pruning performs better on narrower networks. However, further experimentation would be needed to confirm this.

Furthermore, as the size of the network increases, we observe that the accuracy of the winning ticket increases as well. One possible explanation is that as the network size increases, it becomes more likely that an appropriate sub-network (the winning ticket) is found by the pruning methods. However, since we prune a certain percentage of the weights in each case, the size of the winning ticket increases as the size of the initial network increases. This probably also contributes to the increase in accuracy. In the future, it would be interesting to extend this experiment by pruning to a fixed-sized winning ticket irrespective of the initial network size.

\subsection{Over-pruning}

In the previous experiment, we explored how the size of the network influences the performance of different pruning methods. In this experiment, we want to investigate how the pruning methods perform when the network becomes very sparse (up to a 99.6\% reduction in size). As previously, we compare Fisher pruning, L1 unstructured pruning, and random pruning. It is difficult to predict how Fisher pruning and L1 unstructured pruning will perform when pruning the network this severely. However, random pruning is expected to perform worse than both Fisher pruning and L1 unstructured pruning. Furthermore, we suspect that random pruning will result in a drop in accuracy due to the creation of a bottleneck in the network. Such a bottleneck can develop as the random nature of the pruning might result in most (or all) nodes in a layer being removed, which severely limits the information passed to the following layer in the network.

\begin{figure}
     \centering
     \begin{subfigure}[t]{0.49\textwidth}
         \centering
         \includegraphics[width=\textwidth]{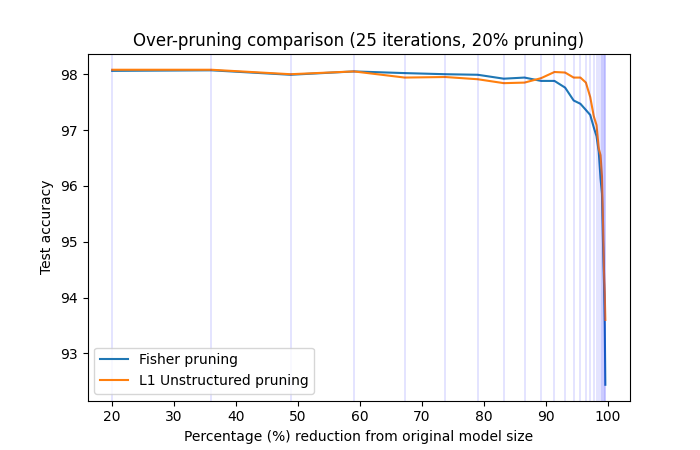}
         \caption{25 pruning iterations}
         \label{fig:over-pruning-1}
     \end{subfigure}
     \hfill
     \begin{subfigure}[t]{0.49\textwidth}
         \centering
         \includegraphics[width=\textwidth]{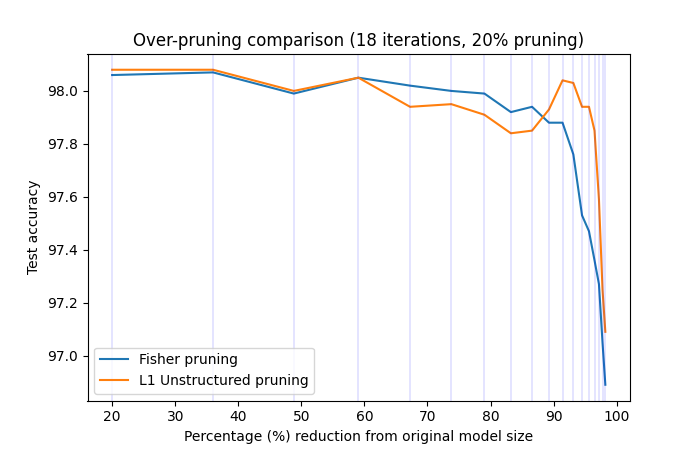}
         \caption{18 pruning iterations}
         \label{fig:over-pruning-2}
     \end{subfigure}
        \caption{Test accuracy of different pruning methods as the percentage of pruned weights is increased (20\% pruning at each iteration).}
        \label{fig:overall-over-pruning-1}
\end{figure}

Figure \ref{fig:overall-over-pruning-1} compares the performance of Fisher pruning and L1 unstructured pruning. Figure \ref{fig:over-pruning-1} shows all 25 pruning iterations (indicated by the vertical lines in the plot), whilst Figure \ref{fig:over-pruning-2} only shows the first 18 pruning iterations. By excluding the last pruning iterations, we are better able to compare the performance as details are harder to make out in Figure \ref{fig:over-pruning-1} due to the final drop in accuracy affecting the scale of the \textit{y}-axis. We observe that the performance of the two pruning methods is initially very similar---for the first 4 pruning iterations, their performance is nearly identical. After this point, some variation occurs, though the performance is still quite similar. This is especially apparent in Figure \ref{fig:over-pruning-1}. However, L1 unstructured pruning results in a small peak in accuracy at around 92\% pruning. Further experimentation would be needed to determine whether this is an anomaly or a feature of L1 unstructured pruning.

\begin{figure}
     \centering
     \begin{subfigure}[t]{0.49\textwidth}
         \centering
         \includegraphics[width=\textwidth]{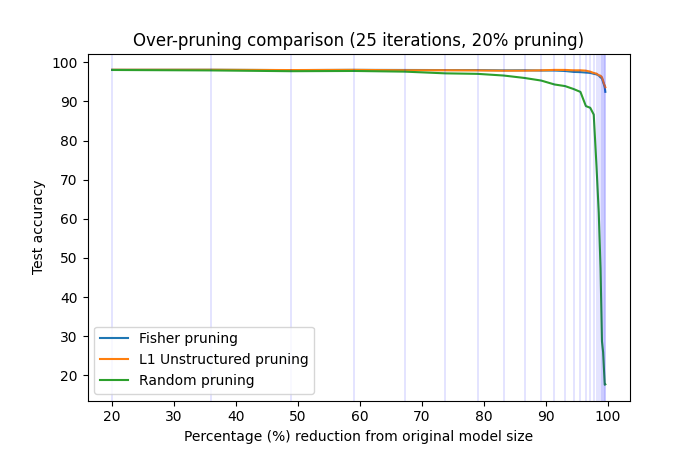}
         \caption{25 pruning iterations}
         \label{fig:over-pruning-3}
     \end{subfigure}
     \hfill
     \begin{subfigure}[t]{0.49\textwidth}
         \centering
         \includegraphics[width=\textwidth]{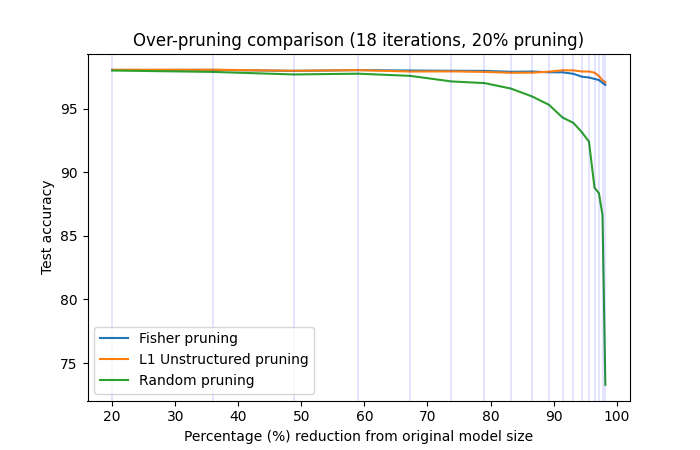}
         \caption{18 pruning iterations}
         \label{fig:over-pruning-4}
     \end{subfigure}
        \caption{Test accuracy of different pruning methods as the percentage of pruned weights is increased (20\% pruning at each iteration).}
        \label{fig:overall-over-pruning-2}
\end{figure}

Figures \ref{fig:over-pruning-3} and \ref{fig:over-pruning-4} are structured similarly to Figures \ref{fig:over-pruning-1} and \ref{fig:over-pruning-2}, but include random pruning. At this scale, the small peak in L1 unstructured pruning becomes increasingly less noticeable. Additionally, our expectations for random pruning are confirmed as it generally performs worse than L1 unstructured pruning and Fisher pruning. After around 68\% pruning, the accuracy starts to steadily decline (see Figure \ref{fig:over-pruning-4}). This could indicate the beginning of a bottleneck developing. However, after around 95\% pruning, the accuracy drops dramatically---going all the way to 20\%. Neither Fisher pruning nor L1 unstructured pruning exhibits a similarly steep drop in accuracy. We suspected that this was due to the aforementioned bottleneck. Since random pruning has no measure of importance for the different weights, it could end up removing nearly all weights from one layer, resulting in a bottleneck. On further inspection, our suspicion was confirmed. An output node in the unpruned network has 100 incoming connections. However, random pruning resulted in each of the output nodes having only 1 or 0 incoming connections. On the other hand, L1 unstructured pruning and Fisher pruning preserved between 21 and 39 incoming connections for the output nodes, allowing the network to make more accuracte predictions.

\subsection{Batched Fisher pruning} \label{section:batched-fisher}

Equation \ref{eq:fisher-pruning} showed that \(\Delta_k\) contains a sum over the squared gradients of all the data points. This means that a backward pass is required for each data point to obtain the gradient for that data point. However, completing one backward pass per data point is very expensive compared to computing a single backward pass based on the combined loss of a batch of data points. Using batches is a standard procedure in machine learning where the batch size is an important (and sensitive) hyperparameter for model performance \cite{batch-size-paper}. Furthermore, if we want to use the gradients that are automatically calculated during model training, then Equation \ref{eq:fisher-pruning} only works if the batch size is set to 1.

For this reason, we decided to investigate the performance of Fisher pruning when using batches of data points. We hypothesised that batches could increase the performance of the pruning method. Furthermore, it would make it easier to incorporate Fisher pruning with the training of standard machine learning models where batches are commonly used. In this case, it would be possible to utilise the gradients calculated during training when computing the Fisher information. When using batches, Equation \ref{eq:fisher-pruning} can be rewritten as Equation \ref{eq:fisher-pruning-batch}. We will refer to this as "batched Fisher pruning".

\begin{equation} \label{eq:fisher-pruning-batch}
    \Delta_k = \frac{1}{2B} \theta^2_k \sum_{b=1}^{B}g^2_{bk}
\end{equation}

where \textit{B} is the number of batches and \(g_{bk}\) is the gradient of the \textit{k}th parameter with respect to the \textit{b}th batch.

For this experiment, we start with a LeNet architecture with two hidden layers of 300 and 100 units respectively, which is trained on the MNIST data set. The models are pruned to approximately 10\% of their original size using iterative Fisher pruning, and the best result from the final training (fine-tuning) of the winning ticket is kept. We repeat the experiment for three different random seeds. For each seed, three batch sizes are tested: 1, 100, and 10,000.

It should be noted that the training procedure of the networks was kept constant throughout the experiment. The default training parameters were used with a batch size of 128. Thus, the Fisher information calculations were performed outside the training loop using a subset (10,000 samples) of the training samples. This allows us to isolate performance differences caused by batched Fisher pruning, rather than performance differences caused by variations in the internal training parameters (such as the training batch size).

\begin{figure}[h]
\centering
\includegraphics[width=0.7\textwidth]{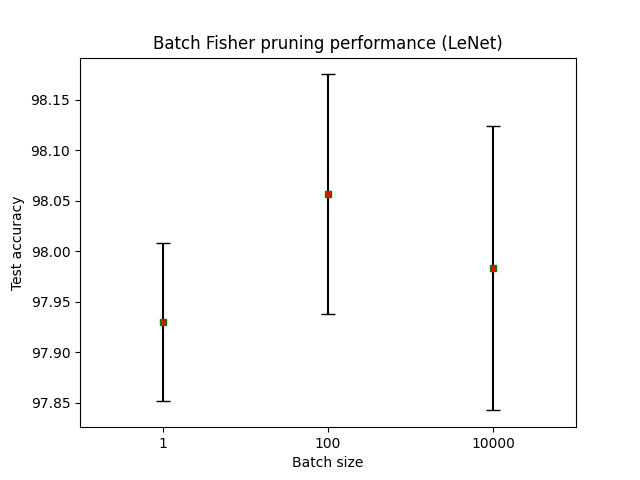}
\caption{Test accuracy of pruned LeNet models (10\% remaining) with different batch sizes for Fisher pruning.}
\label{fig:batch-lenet}
\end{figure}

Figure \ref{fig:batch-lenet} shows the performance of the different batch sizes, where a batch size of 1 is equivalent to the original Fisher pruning algorithm. We observe that increasing the batch size from 1 to 100 results in an increase in the accuracy of the pruned model. Furthermore, with a batch size of 100, we only sum 100 squared gradients instead of 10,000. Further increasing the batch size to 10,000 (i.e., taking the gradient of the loss of all the samples) also seems to increase the accuracy slightly when compared to the original Fisher pruning. However, the standard deviation is big in this case, which might be expected since the gradient is only calculated once.

\begin{figure}[h]
\centering
\includegraphics[width=0.7\textwidth]{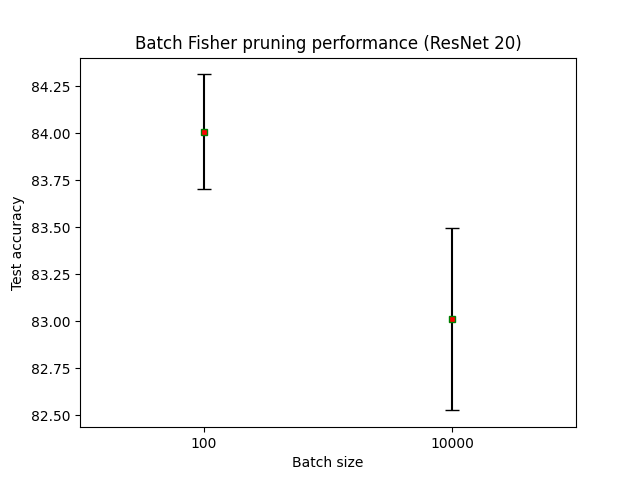}
\caption{Test accuracy of pruned ResNet models (10\% remaining) with different batch sizes for Fisher pruning.}
\label{fig:batch-resnet}
\end{figure}

Since the LeNet architecture is quite simple, we decided to test how batched Fisher pruning performs on a more computationally expensive network. For this, we used the ResNet-20 architecture on the CIFAR-10 data set. The results (see Figure \ref{fig:batch-resnet}) further emphasised the benefit of using batched Fisher pruning. With a batch size of 1, Fisher pruning became unfeasible as it would take approximately a week to finish. This is due to the high cost of performing 10,000 backward passes with such a complex network. However, for the other two batch sizes (100 and 10,000), the training and pruning procedure finished in a couple of hours. In Figure \ref{fig:batch-resnet}, we observe that a higher test accuracy is achieved with a batch size of 100 compared to a batch size of 10,000. A similar trend was seen in Figure \ref{fig:batch-lenet}, however, the difference between the two models is even greater here (Figure \ref{fig:batch-resnet}).

The main focus of this experiment was to investigate the potential benefit, in terms of improved test accuracy, when using batched Fisher pruning. However, batches can improve Fisher pruning in other ways as well. In particular, batched Fisher pruning can lead to a significant increase in speed as it requires fewer backward passes. In this study, we only had time to perform a limited number of tests. In the future, it would be interesting to expand this experiment to gain a better understanding of how batched Fisher pruning can improve performance.

\section{Conclusion}

In this study, we have explored different pruning techniques, with a focus on Fisher pruning and L1 unstructured pruning. Through our experimentation, we have found that one-shot pruning and iterative pruning have very similar performance up to around 50\% reduction in network size, but at high levels of pruning, iterative pruning outperforms one-shot pruning in terms of accuracy. Additionally, we have found that the choice of pruning method affects the amount of weight movement, with L1 unstructured pruning resulting in slightly less weight movement than Fisher pruning. Furthermore, we have found that the performance of the pruning methods varies depending on the width of the network, with Fisher pruning seeming to perform better on wider networks, while L1 unstructured pruning performed better on narrower networks. However, further experimentation would be needed to confirm this. We also found that as the network becomes very sparse, the performance of Fisher pruning and L1 unstructured pruning diverges slightly. However, it would require further investigation to determine whether this is an anomaly or caused by characteristics of the pruning methods. On the other hand, random pruning exhibited a significant drop in accuracy when over-pruning, which is probably due to the creation of a bottleneck in the network. Finally, we have introduced the concept of batched Fisher pruning, which led to an improvement in both test accuracy and computational efficiency when compared to the traditional Fisher pruning method.

\section{Future work}

Our investigation has highlighted several opportunities for future work. One avenue for further exploration is to conduct additional experimentation with varying network widths. This could involve both increasing the number of experiments and expanding the range of networks and data used. Additionally, we proposed pruning networks of different initial sizes down to a fixed-sized winning ticket (rather than a certain percentage). This would allow us to better investigate the effect of larger network sizes since the size of the winning ticket would no longer be proportional to the original network size. Another potential extension is to examine the impact of increased network depth on pruning performance. Furthermore, it would be beneficial to implement batched Fisher pruning with feature map Fisher information, which has been shown to be more appropriate for convolutional models (as suggested in \cite{fisher-pruning}). It would also be interesting to investigate structured pruning, which removes entire neurons instead of single weights, as this constitutes a sort of middle ground between weight pruning and feature map pruning. This could potentially be combined with Fisher information. Lastly, it would be useful to explore the differences in performance between iterative pruning and continuous training methods. 




{
\small
\bibliographystyle{unsrt}
\bibliography{main}
}

\end{document}